\documentclass[a4paper,11pt]{article}

\usepackage[margin=2.5cm]{geometry}

\usepackage[english]{babel}

\usepackage[utf8]{inputenc}

\usepackage[ttscale=.875]{libertine}
\usepackage[scaled=0.96]{zi4}

\usepackage[switch]{lineno}
\usepackage[backend=biber, style=numeric, natbib=true]{biblatex}
\bibliography{draft.bib}

\usepackage{graphicx}

\usepackage{tocloft}
\usepackage{etoc}

\title{A graphical, scalable and intuitive method\\
       for the placement and the connection of biological cells}
\author{Nicolas P. Rougier\footnote{INRIA Bordeaux Sud-Ouest, Talence, France
                                    -- \tt{Nicolas.Rougier@inria.fr}}}
\date{\today}

\begin{document}
\maketitle

\begin{abstract}
  We introduce a graphical method originating from the computer graphics domain
  that is used for the arbitrary and intuitive placement of cells over a
  two-dimensional manifold. Using a bitmap image as input, where the color
  indicates the identity of the different structures and the alpha channel
  indicates the local cell density, this method guarantees a discrete
  distribution of cell position respecting the local density function. This
  method scales to any number of cells, allows to specify several different
  structures at once with arbitrary shapes and provides a scalable and
  versatile alternative to the more classical assumption of a uniform
  non-spatial distribution. Furthermore, several connection schemes can be
  derived from the paired distances between cells using either an automatic
  mapping or a user-defined local reference frame, providing new computational
  properties for the underlying model. The method is illustrated on a discrete
  homogeneous neural field, on the distribution of cones and rods in the retina
  and on a coronal view of the basal ganglia.\\

\end{abstract}

\renewcommand\contentsname{}
\tableofcontents
\clearpage

% ------------------------------------------------------------ Introduction ---
\section{Introduction}

\begin{figure}
  \includegraphics[width=\textwidth]{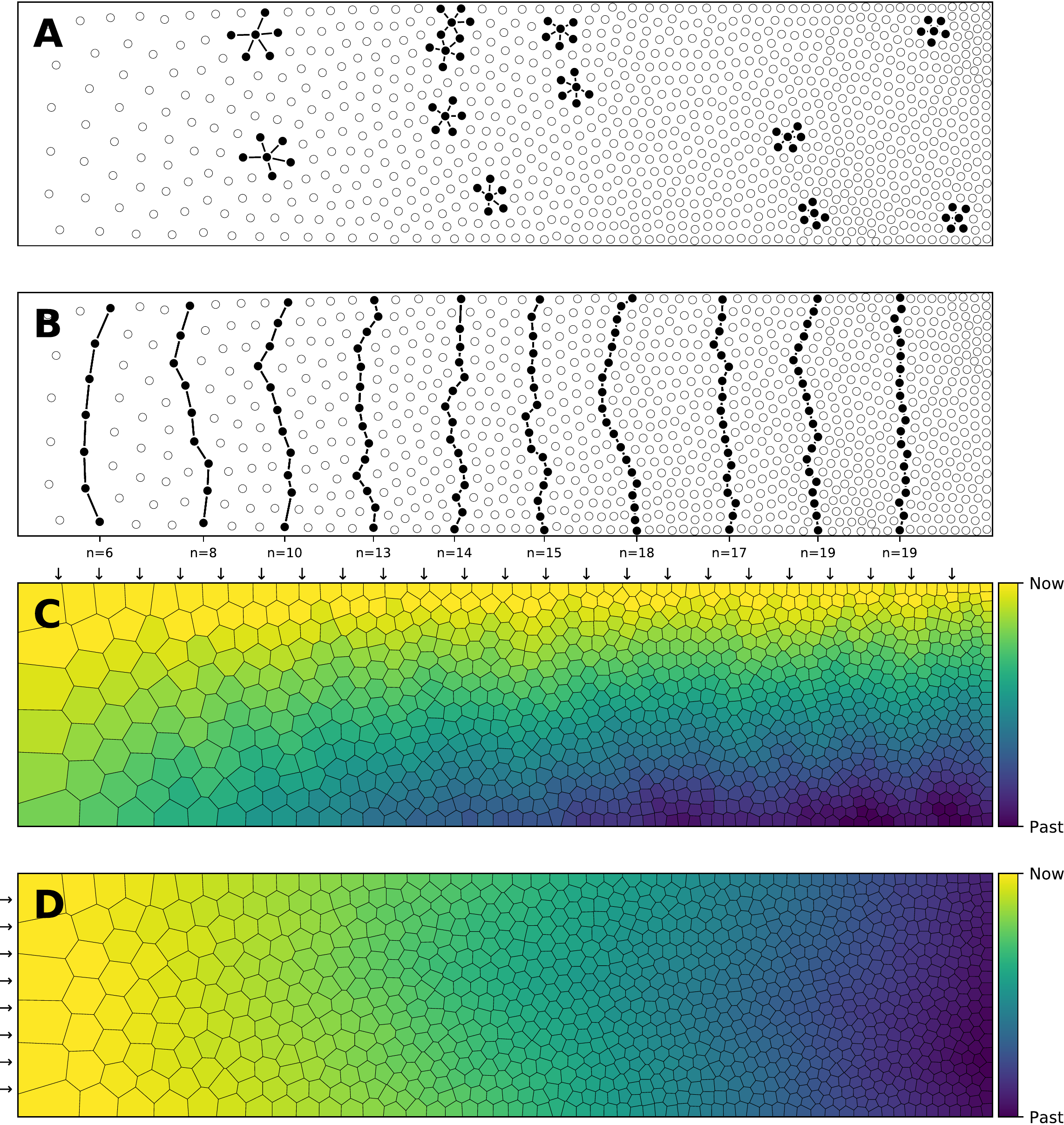}  
  \caption{\textbf{Influence of spatial distribution on signal propagation.}
    \textbf{\textsf{A.}} A k-nearest neighbours (k=5) connectivity pattern
    shows mid-range connection lengths in low local density areas (left part)
    and short-range connection lengths in high density areas (right
    part). \textbf{\textsf{B.}} Shortest path from top to bottom using a
    k-nearest neighbours connectivity pattern (k=5). The lower the density, the
    shorter the path and the higher the density, the longer the path. On the
    far left, the shortest path from top to bottom is only 6 connections while
    this size triples on the far right to reach 19 connections. Said
    differently, the left part is the fast pathway while the right part is the
    slow pathway relatively to some input data that would feed the architecture
    from the top. \textbf{\textsf{C.}} Due to the asymmetry of cells position,
    a signal entering on the top side (materialized with small arrows) travels
    at different speeds and will consequently reach the bottom side at
    different times. This represents a spatialization of
    time. \textbf{\textsf{D.}} Due to the asymmetry of cells position, a signal
    entering on the left side (materialized with small arrows) slows down while
    traveling before reaching the right side. This represents a compression of
    time and may serve as a short-term working memory.}
  \label{fig:diffusion}
\end{figure}

The spatial localization of neurons in the brain plays a critical role since
their connectivity patterns largely depends on their type and their position
relatively to nearby neurons and regions (short-range or/and long-range
connections). Interestingly enough, if the neuroscience literature provides
many data about the spatial distribution of neurons in different areas and
species (e.g. \cite{Pasternak:1975} about the spatial distribution of neurons
in the mouse barrel cortex, \cite{McCormick:2000} about the neuron spatial
distribution and morphology in the human cortex, \cite{Blazquez-Llorca:2014}
about the spatial distribution of neurons innervated by chandelier cells), the
computational literature exploiting such data is rather scarce and the spatial
localization is hardly taken into account in most neural network models (be it
computational, cognitive or machine learning models). One reason may be the
inherent difficulty in describing the precise topography of a population such
that most of the time, only the overall topology is described in term of
layers, structures or groups with their associated connectivity patterns (one
to one, one to all, receptive fields, etc.). One can also argue that such
precise localization is not necessary because for some model, it is not
relevant (machine learning) while for some others, it may be subsumed into the
notion of cell assemblies \cite{Hebb:1949} that represent the spatiotemporal
structure of a group of neurons wired and acting together. Considering cell
assemblies as the basic computational unit, one can consider there is actually
few or no interaction between assemblies of the same group and consequently,
their spatial position is not relevant. However, if cell assemblies allows to
greatly simplify models, it also brings implicit limitations whose some have
been highlighted in \cite{Nallapu:2017}. To overcome such limitations, we
think the spatial localization of neurons is an important criterion worth to be
studied because it could induces original connectivity schemes from which new
computational properties can be derived as it is illustrated on figure
\ref{fig:diffusion}.\\
\begin{figure}[htbp]
  \includegraphics[width=.5\textwidth]{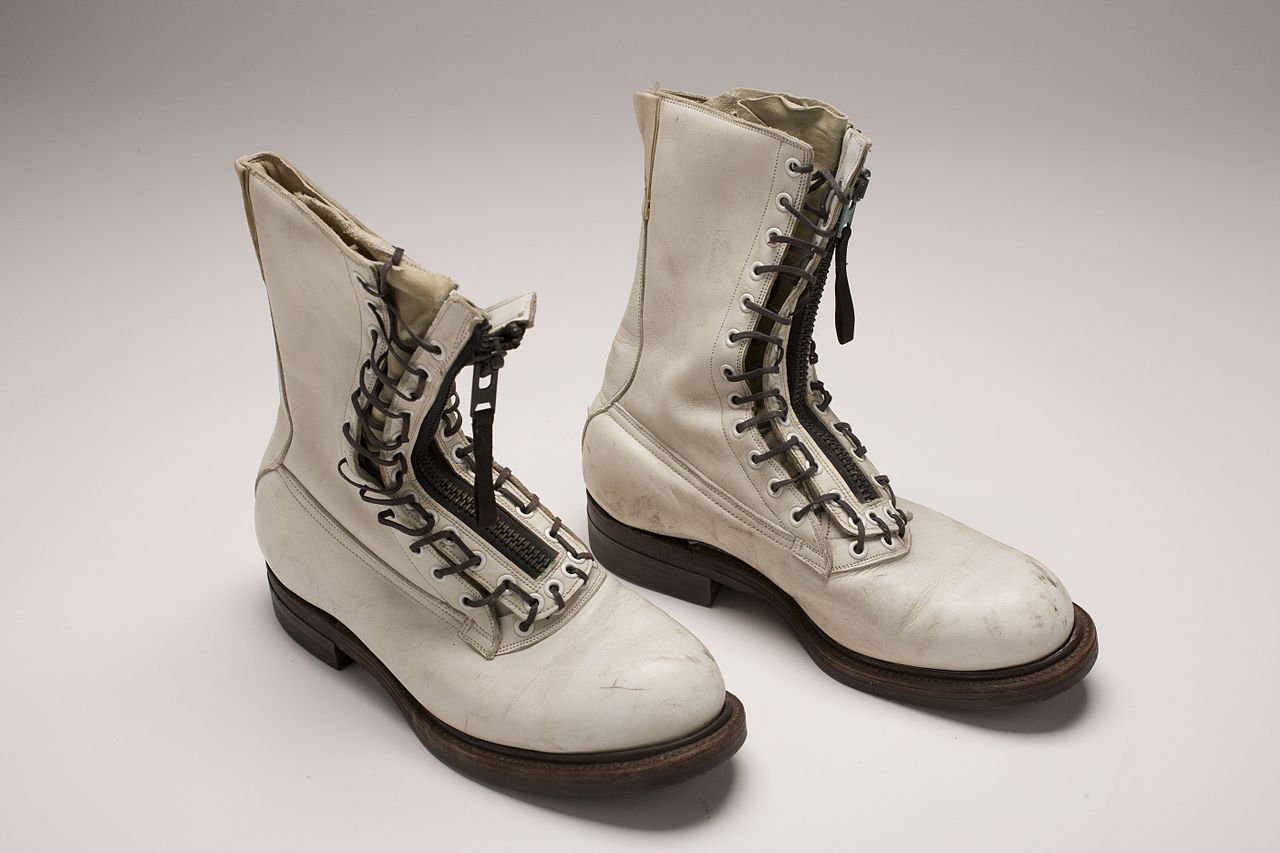}
  \includegraphics[width=.5\textwidth]{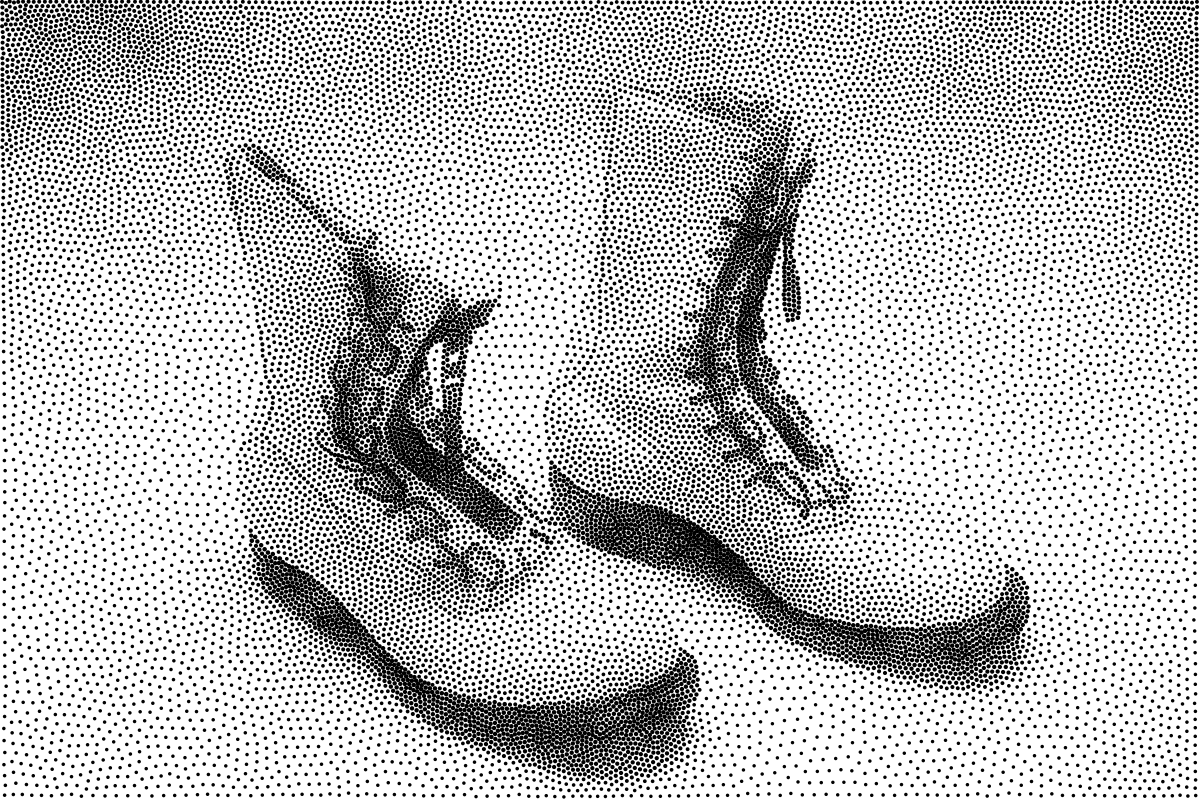}
  \caption{\textbf{Stippling.} According to
    Wikipedia\protect\footnotemark, {\em Stippling is the creation of a pattern
      simulating varying degrees of solidity or shading by using small
      dots. Such a pattern may occur in nature and these effects are frequently
      emulated by artists.} The pair of boots (left part) have been first
    converted into a gray-level image and processed into a stippling figure
    (right part) using the weighted Voronoi stippling technique by
    \cite{Secord:2002} and replicated in \cite{Rougier:2017}. Image from
    \cite{Rougier:2017} (CC-BY license).}
  \label{fig:boots}
\end{figure}

However, before studying the influence of the spatial localisation of neurons,
it is necessary to design first a method for the arbitrary placement of
neurons. This article introduces a graphical, scalable and intuitive method for
the placement of neurons (or any other type of cells actually) over a
two-dimensional manifold and provides as well the necessary information to
connect neurons together using either an automatic mapping or a user-defined
function. This graphical method is based on a stippling techniques originating
from the computer graphics domain for non-photorealistic rendering as
illustrated on figure \ref{fig:boots}.

\footnotetext{
  Stippling Wikipedia entry at {\tt https://en.wikipedia.org/wiki/Stippling}}

% ----------------------------------------------------------------- Methods ---
\section{Methods}

\begin{figure}[htbp]
  \includegraphics[width=\textwidth]{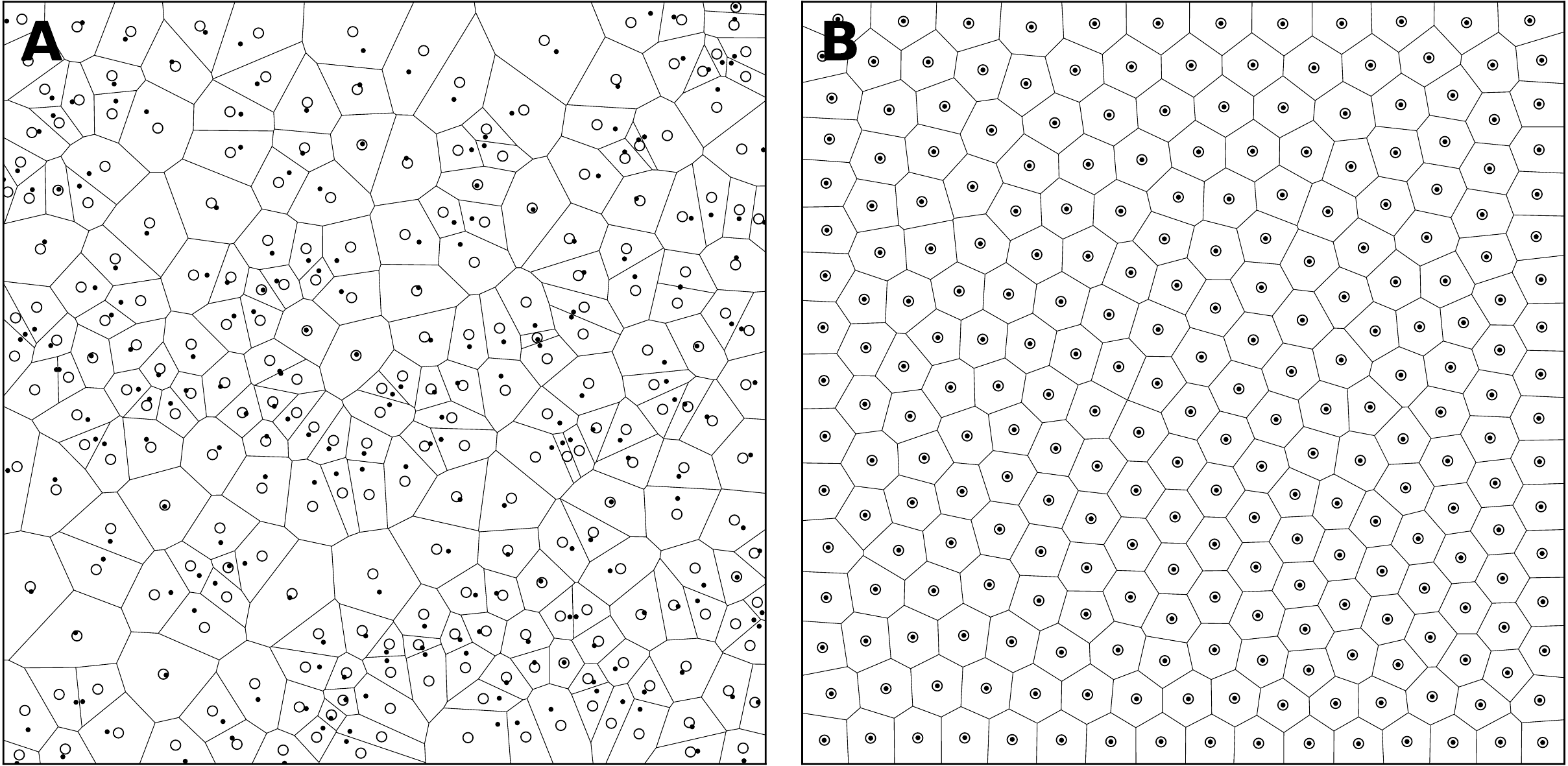}
  \caption{\textbf{Centroidal Voronoi Tesselation.}  \textbf{\textsf{A.}}
    Voronoi diagram of a uniform distribution (n=256) where black dots
    represent the uniform distribution and white circles represent the
    centroids of each Voronoi cells. \textbf{\textsf{B.}} Centroidal Voronoi
    diagram where the point distribution matches the centroid distribution.}
  \label{fig:CVT}
\end{figure}

Blue noise \cite{Ulichney:1987} is {\em an even, isotropic yet unstructured
  distribution of points} \cite{Mehta:2012} and has {\em minimal low frequency
  components and no concentrated spikes in the power spectrum energy}
\cite{Zhang:2016}. Said differently, blue noise (in the spatial domain) is a
type of noise with intuitively good properties: points are evenly spread
without visible structure (see figure \ref{fig:CVT} for the comparison of a
uniform distribution and a blue noise distribution). This kind of noise has been
extensively studied in the computer graphic domain and image processing because
it can be used for object distribution, sampling, printing, half-toning,
etc. One specific type of spatial blue noise is the Poisson disc distribution
that is a 2D uniform point distribution in which all points are separated from
each other by a minimum radius (see right part of figure
\ref{fig:CVT}). Several methods have been proposed for the generation of such
noise, from the best in quality (dart throwing \cite{Cook:1986}) to faster ones
(rejection sampling \cite{Bridson:2007}), see \cite{Lagae:2008} for a
review. An interesting variant of the Poisson disk distribution is a non
isotropic distribution where local variations follow a given density function
as illustrated on figure \ref{fig:boots} where the density function has been
specified using the image gray levels. On the stippling image on the right,
darker areas have a high concentration of dots (e.g. boots sole) while lighter
areas such as the background display a sparse number of dots. There exist
several techniques for computing such stippling density-driven pattern (optimal
transport \cite{Mehta:2012}, variational approach \cite{Chen:2012}, least
squares quantization \cite{Lloyd:1982}, etc.) but the one by \cite{Secord:2002}
is probably the most straightforward and simple and has been recently replicated
in \cite{Rougier:2017}.

\subsection{Distribution}

\begin{figure}
  \includegraphics[width=\textwidth]{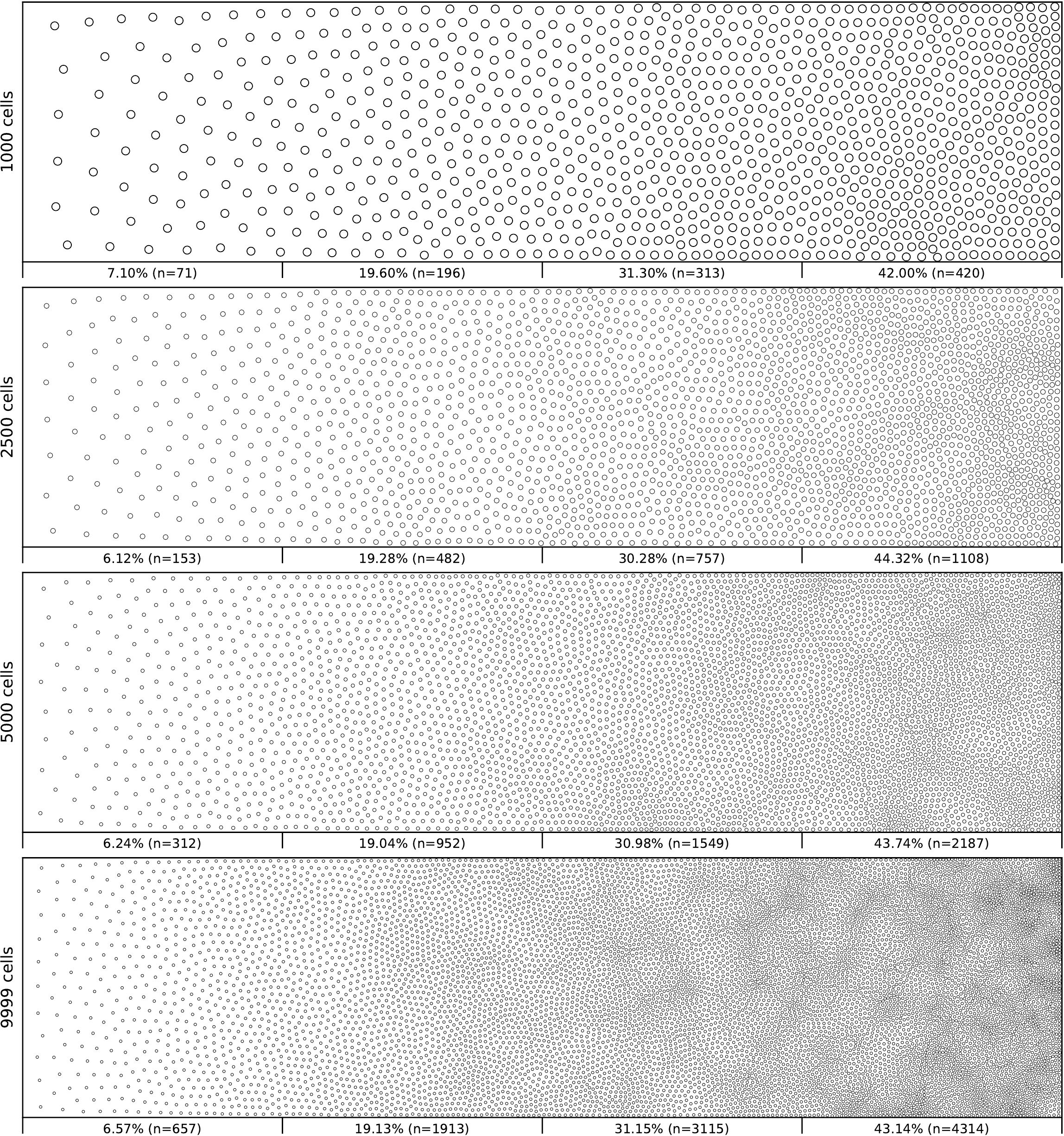}  
  \caption{\textbf{Non-uniform distribution (linear gradient).} Different
    population distribution (size of 1000, 2500, 5000 and 10000 cells) using
    the same linear gradient as input have been computed. Each distribution has
    been split into four equal areas and the respective proportion and number
    of cells present in the area is indicated at the bottom of the area. The
    proportion of cells present in each areas is approximately independent
    ($\pm$2.5\%) of the overall number of cells. }
  \label{fig:gradient}
\end{figure}

The desired distribution is given through a bitmap RGBA image that provides two
types of information. The three color channels indicates the identity of a cell
(using a simple formula of the type $identity = 256 \times 256 \times R + 256
\times G + B$ for $0 \leq R,G,B < 256$) and the alpha channel indicates the
desired local density. This input bitmap has first to be resized (without
interpolation) such that the mean pixel area of a Voronoi cell is 500
pixels. For example, if we want a final number of 1000 cells, the input image
needs to be resized such that it contains at least 500x1000 pixels. For
computing the weighted centroid, we apply the definition proposed in
\cite{Secord:2002} over the discrete representation of the domain and use a
LLoyd relaxation scheme.
\[
  {\bf C}_i = \frac{\int_A {\bf x}\rho({\bf x})dA}{\int_A \rho({\bf x})}
\]
More precisely, each Voronoi cell is rasterized (as a set of pixels) and the
centroid is computed (using the optimization proposed by the author that allow
to avoid to compute the integrals over the whole set of pixels composing the
Voronoi cell). As noted by the author, the precision of the method is directly
related to the size of the Voronoi cell. Consequently, if the original density
image is too small relatively to the number of cells, there might be quality
issues. We use a fixed number of iterations ($n=50$) instead of using the
difference in the standard deviation of the area of the Voronoi regions as
proposed in the original paper. Last, we added a threshold parameter that
allows to perform a pre-processing of the density image: any pixel with an
alpha level above the threshold is set to the threshold value before
normalizing the alpha channel. Figure \ref{fig:gradient} shows the distribution
of four populations with respective size 1000, 2500, 5000 and 10000 cells,
using the same linear gradient as input. It is remarkable to see that the local
density is approximately independent of the total number of cells.

\subsection{Connection}

\begin{figure}
  \includegraphics[width=\textwidth]{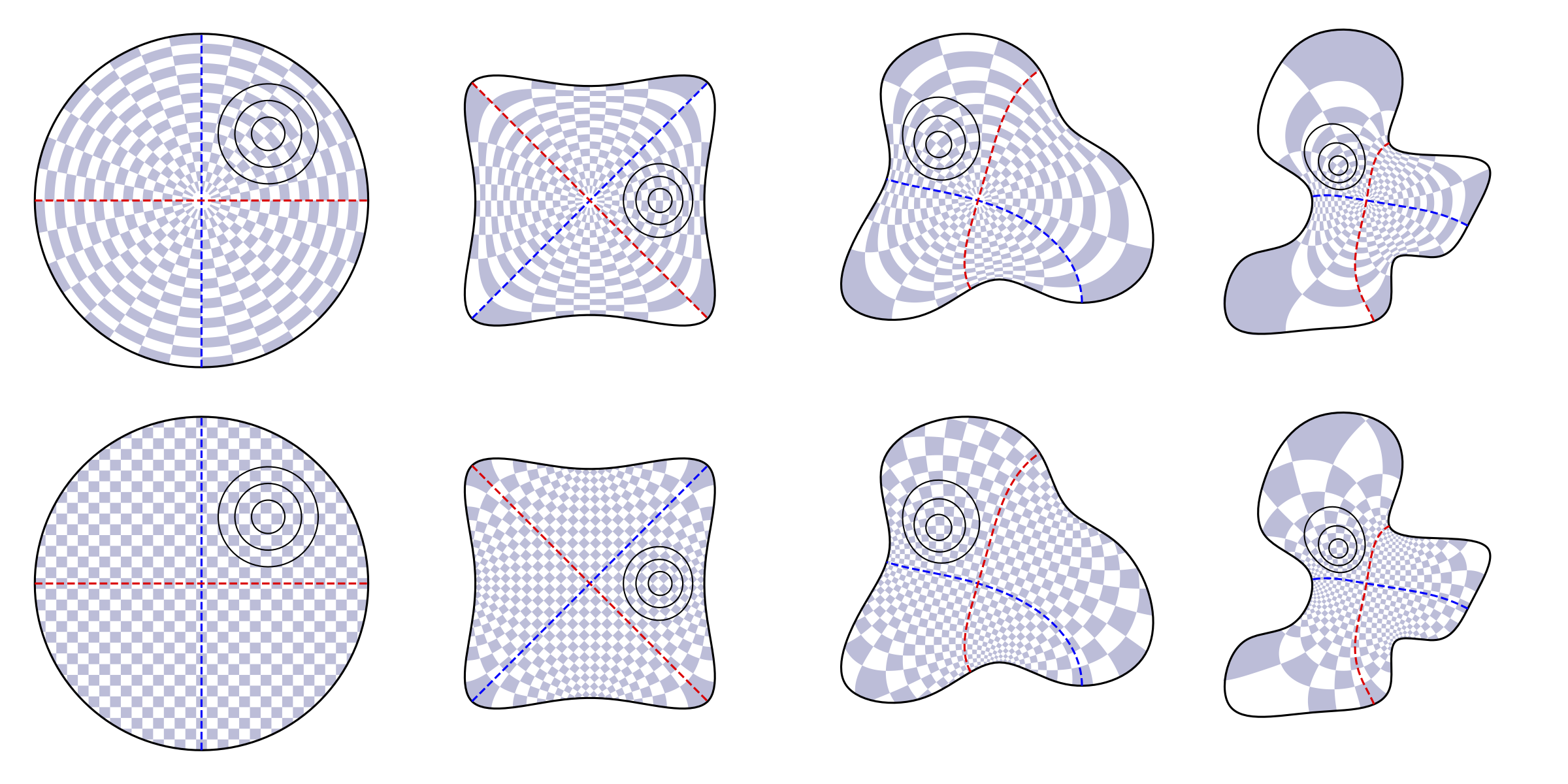}  
  \caption{\textbf{Conformal mappings.} Examples of conformal mappings on
    arbitrary spline shapes using the conformal Riemann mapping via the Szegö
    kernel \cite{Kerzman:1986}. Top line shows conformal mapping of the polar
    domain, bottom line show conformal mapping of the Cartesian domain.}
  \label{fig:mapping}
\end{figure}

Most computational models need to define the connectivity between the different
populations that compose the model. This can be done by specifying projections
between a source population and a target population. Such projections
correspond to the axon of the source neuron making a synaptic contact with the
dendritic tree of the target neuron. In order to define the overall model
connectivity, one can specify each individual projection if the model is small
enough (a few neurons). However, for larger models (hundreds, thousands or
millions of neurons), this individual specification would be too cumbersome and
would hide any structure in the connectivity scheme. Instead, one can use
generic connectivity description \cite{Djurfeldt:2014} such as one-to-one,
one-to-all, convergent, divergent, receptive fields, convolutional, etc. For
such connectivity scheme to be enforced, it requires either a well structured
populations (e.g. grid) or a simple enclosing topology \cite{Ekkehard:2015}
such as a rectangle or a disc. In the case of arbitrary shapes as shown on
figure \ref{fig:mapping}, these methods cannot be used directly. However, we
can use an indirect mapping from a reference shape such as the unit disc and
take advantage of the Riemann mapping theorem that states (definition from
\cite{Bolt:2010}):\\

\textbf{Riemann mapping theorem} (from \cite{Bolt:2010}). {\em Let $\Omega$ be
  a (non empty) simply connected region in the complex plane that is not the
  entire plane. Then, for any $z_0 \in \Omega$, there exists a bianalytic
  (i.e. biholomorphic) map $f$ from $\Omega$ to the unit disc such that
  $f(z0)=0$ and $f'(z0)>0$.}\\

Such mapping is {\em conformal}, that it, it preserves angles while {\em
  isometric} mapping preserves lengths (developable surfaces) and {\em
  equiareal} mapping preserves areas. \citet{Kerzman:1986} introduced a method
to compute the Riemann mapping function using the Szegö kernel that is
numerically stable while \citet{Trefethen:1980} introduced numerical methods
for solving the more specific conformal Schwarz-Christoffel transformation
(conformal transformation of the upper half-plane onto the interior of a simple
polygon). Furthermore, a Matlab toolkit is available in \cite{Driscoll:1996} as
well as a Python translation (\url{https://github.com/AndrewWalker/cmtoolkit})
that has been used to produce the figure \ref{fig:mapping} that shows some examples of
arbitrary shapes and the automatic mapping of the polar and Cartesian domains.
However, even if automatic, this mapping can be perceived as not
intuitive. Provided the shape are not too distorted, we'll see in the results
section that ad-hoc mapping can also be used.

\subsection{Visualization}

Having now a precise localization for each cell of each population, we have
several ways of visualizing the activity within the model. The most
straightforward way is to simply draw the activity of a cell at its position
using a disc of varying color (a.k.a. colormap) or varying size, correlated
with cell activity. This requires the total number of cells to be not too large
or the display would be cluttered. For a moderate number of cells, we can take
advantage of the dual Voronoi diagram of the cell position as illustrated on
figure \ref{fig:diffusion}, using a colormap to paint the Voronoi
cell. Finally, if the number of cells is really high, A two-dimensional
histogram of the mean activity (with a fixed number of bins) can be used as
shown on figure \ref{fig:BG}C using a bicubic interpolation filter.

% ----------------------------------------------------------------- Results ---
\section{Results}

We'll now illustrate the use of the proposed method on three different cases.

\subsection{Case 1: Retina cells}

\begin{figure}
  \includegraphics[width=\textwidth]{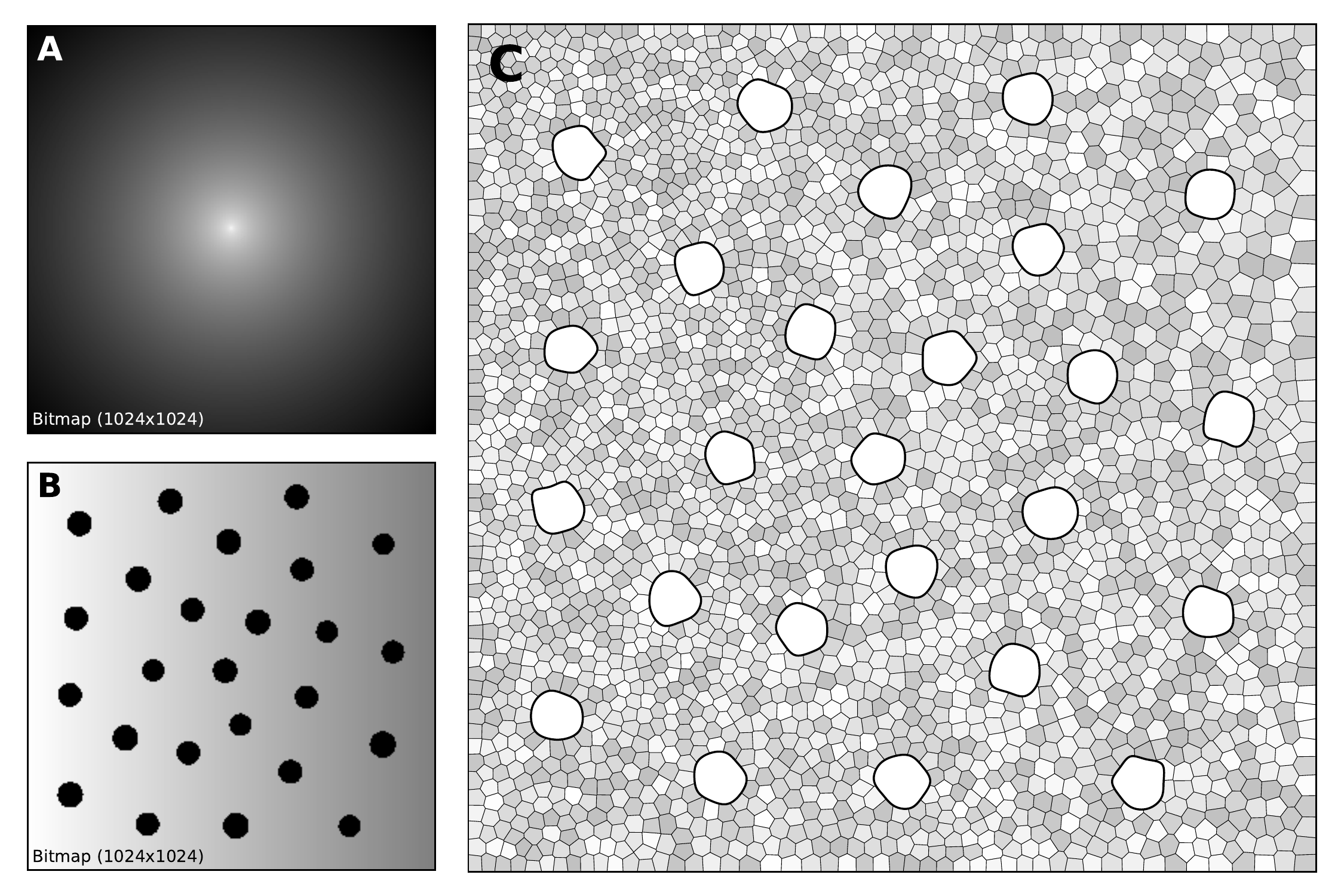}  
  \caption{\textbf{Cones and rods distribution.}  \textbf{\textsf{A.}} The
    density map for cones placement (n=25) is a circular and quadratic gradient
    with highest density in the center. \textbf{\textsf{B.}} The density map
    for rods placement (n=2500) is built using the rods distribution. Starting
    from a linear density, ``holes'' with different sized are created at the
    location of each cone, preventing rods to spread over these areas during
    the stippling procedure. \textbf{\textsf{C.}}  Final distribution of cones
    and rods. Cones are represented as white blobs (splines) while rods are
    represented as Voronoi regions using random colors to better highlight the
    covered area.}
  \label{fig:retina}
\end{figure}

The human retina counts two main types of photoreceptors, namely rods, and
cones (L-cones, M-cones and S-cones). They are distributed over the retinal
surface in an non uniform way, with a high concentration of cones (L-cones and
M-cones) in the foveal region while the rods are to be found mostly in the
peripheral region with a peak density at around 18-20$^\circ$ of foveal
eccentricity. Furthermore, the respective size of those cells is different,
rods being much smaller than cones. The distribution of rods and cells in the
human retina has been extensively studied in the literature and is described
precisely in a number of work \cite{Curcio:1990,Ahnelt:2000}. Our goal here is
not to fit the precise distribution of cones and rods but rather to give a
generic procedure that can be eventually used to fit those figures, for a
specific region of the retina or the whole retina. The main difficulty is the
presence of two types of cells having different sizes. Even though there exist
blue-noise sampling procedures taking different size into account
\cite{Zhang:2016}, we'll use instead the aforementioned method using a two
stage procedure as illustrated on figure \ref{fig:retina}.

A first radial density map is created for the placement of 25 cones and the
stippling procedure is applied for 15 steps to get the final positions of the 25
cones. A linear rod density map is created where discs of varying (random)
sizes of null density are created at the position of the cones. These discs will
prevent the rods to spread over these areas. Finally, the stippling procedure
is applied a second time over the built density map for 25 iterations. The
final result can be seen on figure \ref{fig:retina}C where rods are tightly
packed on the left, loosely packed on the left and nicely circumvent the cones.

\subsection{Case 2: Neural field}

\begin{figure}[htbp]
  \begin{center}
  \includegraphics[width=.32\textwidth]{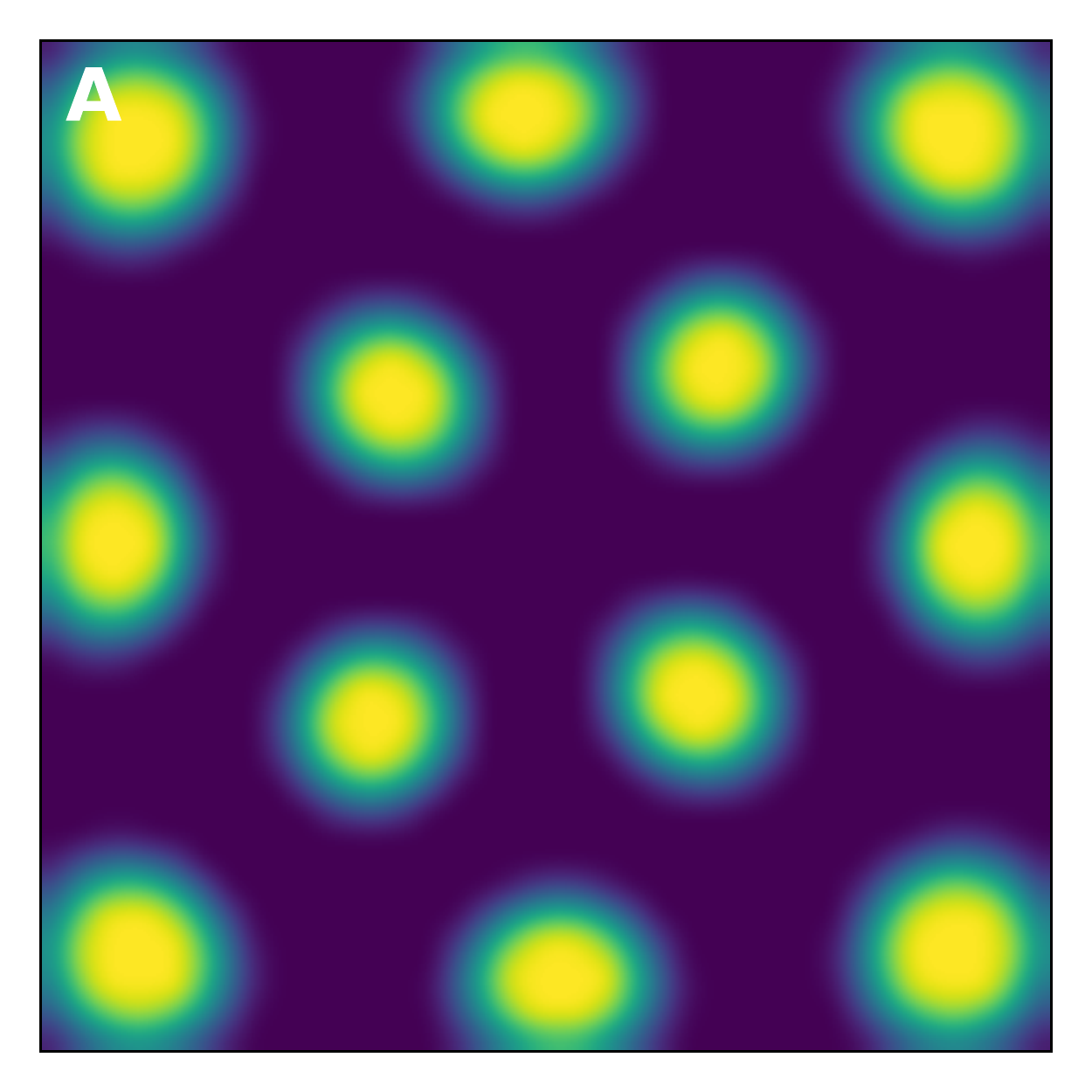}
  \includegraphics[width=.32\textwidth]{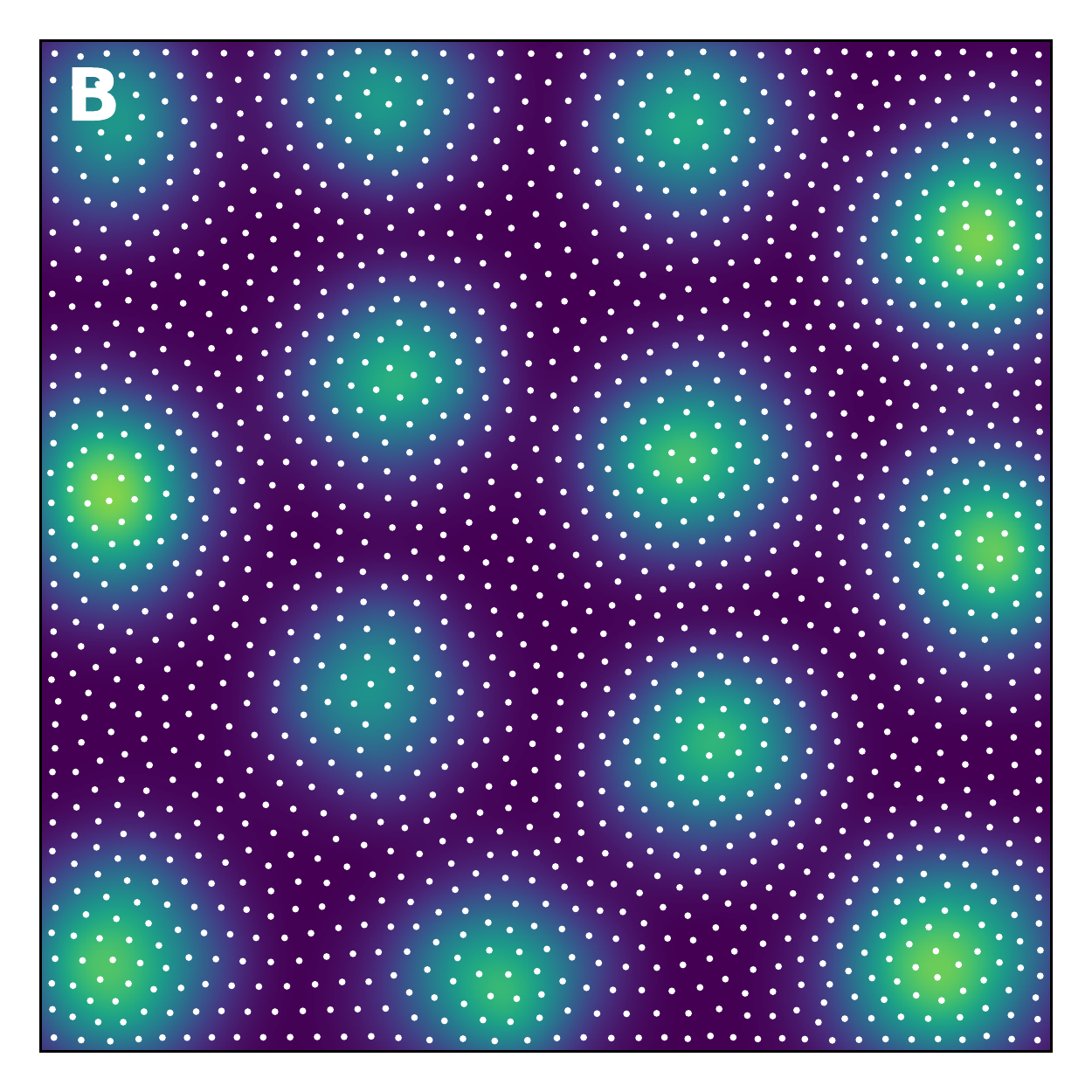}
  \includegraphics[width=.32\textwidth]{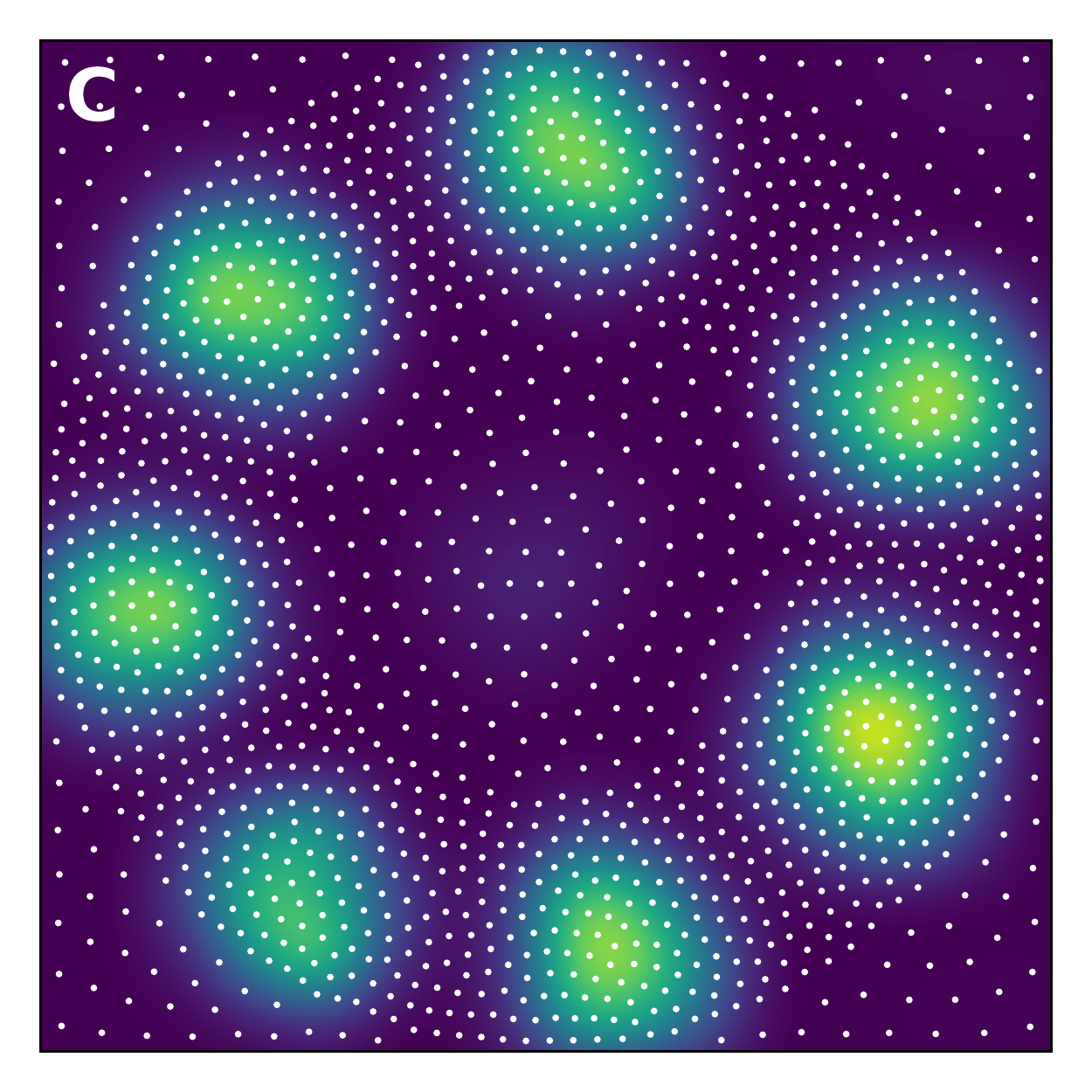}
  \end{center}
  \caption{\textbf{Non-homogeneous discrete neural field}.
    \textbf{\textsf{A.}}  Turing patterns resulting from a continuous and
    homogeneous neural field with constant and noisy
    input. \textbf{\textsf{B.}} Turing patterns resulting from a discrete and
    homogeneous neural field with constant and noisy input. White dots indicate
    the position of the cells. Mean activity is compute from the histogram of
    cells activity using $40 \times 40$ bins. \textbf{\textsf{C.}} Localized
    Turing patterns resulting from a discrete and non-homogeneous neural field
    with constant and noisy input. White dots indicate the position of the
    cells. Mean activity is computed from the histogram of cells activity using
    $40 \times 40$ bins. }
  \label{fig:DNF}
\end{figure}

Neural fields describe the dynamics of a large population of neurons by taking
the continuum limit in space, using coarse-grained properties of single neurons
to describe the activity
\cite{Wilson:1972,Wilson:1973,Amari:1977,Coombes:2014}. In this example, we
consider a neural field with activity $u$ that is governed by an equation of
the type:
\[
\tau\frac{\partial u(x,t)}{\partial t} = -u(x,t) + \int_{-\infty}^{+\infty} w(x,y) f(u(y,t)) dy + I(x) + h
\]
The lateral connection kernel $w$ is a difference of Gaussian (DoG) with short
range excitation and long range inhibition and the input $I(x)$ is constant
and noisy. In order to solve the neural field equation, the spatial domain has
been discretized into $40 \times 40$ cells and the temporal resolution has been
set to $10ms$. On figure \ref{fig:DNF}A, one can see the characteristic Turing
patterns that have formed within the field. The number and size of clusters
depends on the lateral connection kernel. Figure \ref{fig:DNF}B shows the
discretized and homogeneous version of the DNF where each cell has been assigned
a position on the field, the connection kernel function and the parameters
being the same as in the continuous version. The result of the simulation shown
on figure \ref{fig:DNF}B is the histogram of cell activities using $40 \times
40$ regular bins. One can see the formation of the Turing patterns that are
similar to the continuous version. On figure \ref{fig:DNF}C however, the
position of the cells have been changed (using the proposed stippling method)
such that there is a torus of higher density. This is the only difference with
the previous model. While the output can still be considered to be Turing
patterns, one can see clearly that the activity clusters are precisely
localized onto the higher density regions. Said differently, the functional
properties of the field have been modified by a mere change in the
structure. This tends to suggest that the homogeneous condition of neural fields
(that is the standard hypothesis in most works because it facilitates the
mathematical study) is actually quite a strong limitation that constrains the
functional properties of the field.

%\subsubsection{Homogenous distribution}
%\subsubsection{Non-homogenous distribution}

%\subsubsection{Cones distribution}
%\subsubsection{Rods distribution}

\subsection{Case 3: Basal ganglia}

\begin{figure}
  \includegraphics[width=\textwidth]{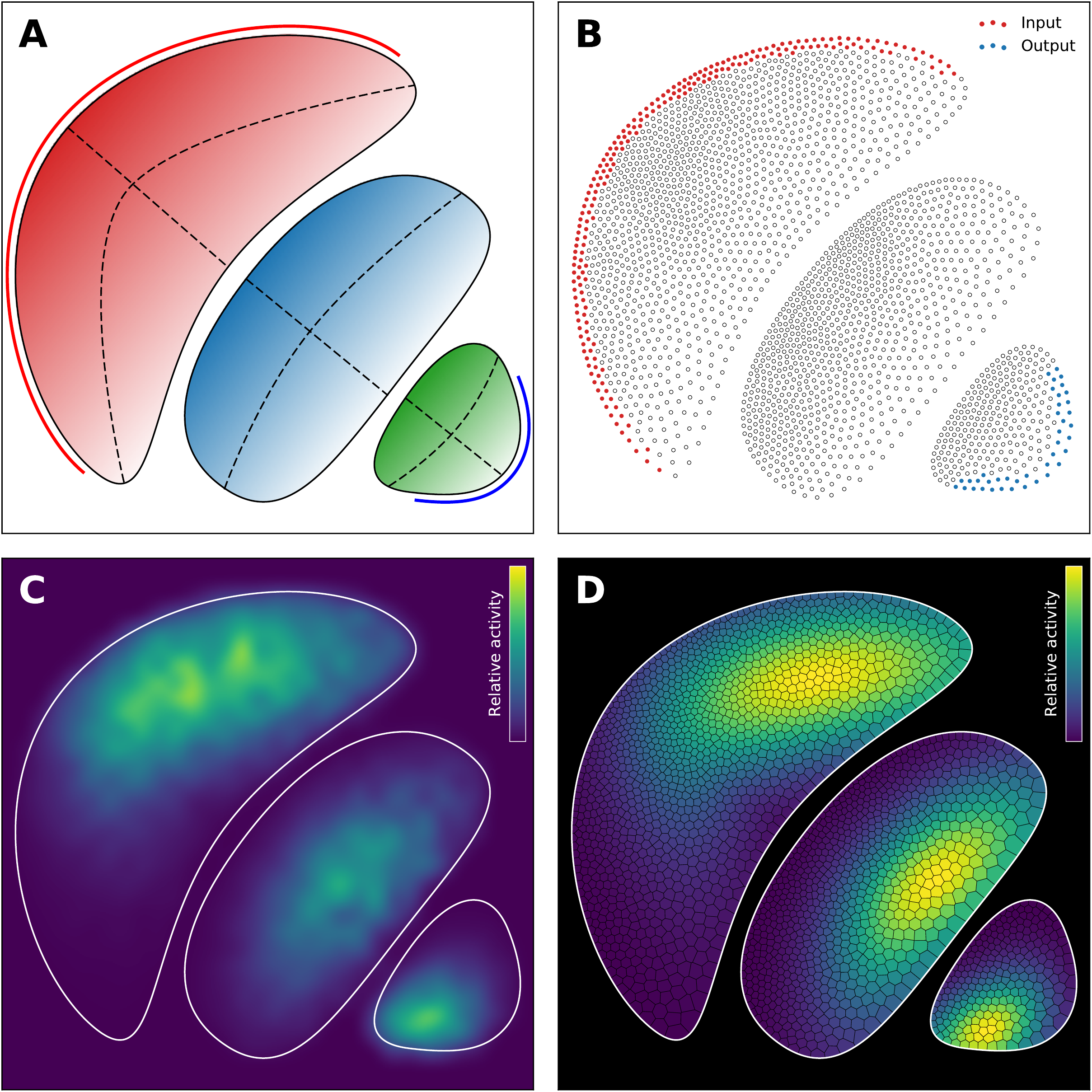}  
  \caption{\textbf{Coronal view of the basal ganglia.} \textbf{\textsf{A.}}
    Scalable Vector Graphic (SVG) source file defining each structure in terms
    of border (solid black lines), major and minor axis (dashed lines), input
    (red line) and output (blue line). Local density is given by the alpha
    channel and structure identity is given by the color. In this coronal view
    of the basal ganglia, the Caudate is red (RGB=(0.83,0.15,0.15)), the GPe is
    blue (RGB=(0.12,0.46,0.70)) and the GPi is green (RGB=(0.17,0.62,0.17)).
    \textbf{\textsf{B.}}  Distribution of 2500 neurons respecting the local
    density and structural organization (Caudate: 1345 cells, GPe: 884 cells,
    GPi: 271 cells). Neurons receiving input are drawn in red, neurons sending
    output are drawn in blue.  Each neuron possesses two set of coordinates:
    one global Cartesian coordinate set and a local curvilinear coordinates set
    defined as the distances to the major and the minor axis of the structure
    the neuron belongs to. \textbf{\textsf{C.}}  Mean activity histogram of the
    different structures using 32x32 bins and a bi-cubic interpolation
    filter. Each bin includes from zero to several neurons. \textbf{\textsf{D.}}
    Cell activities represented using the dual Voronoi diagram of the cell
    position. Each Voronoi region is painted according to the activity of the
    corresponding centroid (i.e. neuron). }
    \label{fig:BG}
\end{figure}

The basal ganglia is a group of sub-cortical nuclei (striatum, globus pallidus,
subthamalic nucleus, subtantia nigra) associated with several functions such as
motor control, action selection and decision making. There exists a functional
dissociation of the ventral and the dorsal part of the striatum (caudate,
putamen and nucleus accumbens) that is believed to play an important role in
decision making \cite{ODoherty:2004,Balleine:2007,Meer:2011} since these two
regions do not receive input from the same structures. For a number of models,
this functional dissociation results in the dissociation of the striatum into
two distinct neural groups even though such anatomical dissociation does not
exist {\em per se} (see \cite{Humphries:2010}). Without any proper topography
of the striatal nucleus, it is probably the most straightforward way to
proceed. However, if each group would possess its own topography, it would become
possible to distinguish the ventral from the dorsal part of the BG, as
illustrated on figure \ref{fig:BG} on a coronal view of the BG. We do not
pretend this simplified view is sufficient to give account on all the intricate
connections between the different nuclei composing the basal ganglia, but it
might nonetheless help to have better understanding of the structure because it
becomes possible to link external input to specific part of this or that
structure (eg. ventral or dorsal part of the striatum). This could lead to
differential processing in different part of the striatum and may reconcile
different theories regarding the role of the ventral and the dorsal part.

%\subsubsection{Cells distribution}
%\subsubsection{Parallel connections}

%\section{Results}
% \subsection{Non homogenous discrete neural field}
% \subsection{Generalized multi-layer perceptron}

% -------------------------------------------------------------- Discussion ---
\section{Discussion}

We've introduced a graphical, scalable and intuitive method for the placement
and the connection of biological cells and we illustrated its use on three
use-cases. We believe this method, even if simple and obvious, might be
worth to be considered in the design of a new class of model, in between
symbolic model and realistic model. Our intuition is that such topography may
be an important aspect that needs to be taken into account and studied in order
for the model to benefit from structural functionality. Furthermore, the
proposed specification of the architecture as an SVG file associated with the
scalability of the method could guarantee to some extent the scalability of the
properties of the model.\\

\textbf{Notes:} All figures were produced using the Python scientific stack,
namely, SciPy \cite{Jones:2001}, Matplotlib \cite{Hunter:2007} and NumPy
\cite{Walt:2011}. All sources are available on GitHub \cite{rougier:2017b}

% -------------------------------------------------------------- References ---
\renewcommand*{\bibfont}{\small}
\printbibliography[title=References]

\end{document}